# Mutual Reinforcement Effects in Japanese Sentence Classification and Named Entity Recognition Tasks

Chengguang Gan†, Qinghao Zhang††, and Tatsunori Mori†

Information extraction(IE) is a crucial subfield within natural language processing. However, for the traditionally segmented approach to sentence classification and Named Entity Recognition, the intricate interactions between these individual subtasks remain largely uninvestigated. In this study, we propose an integrative analysis, converging sentence classification with Named Entity Recognition, with the objective to unveil and comprehend the mutual reinforcement effect within these two information extraction subtasks. To achieve this, we introduce a Sentence Classification and Named Entity Recognition Multi-task (SCNM) approach that combines Sentence Classification (SC) and Named Entity Recognition (NER). We develop a Sentence-to-Label Generation (SLG) framework for SCNM and construct a Wikipedia dataset containing both SC and NER. Using a format converter, we unify input formats and employ a generative model to generate SC-labels, NER-labels, and associated text segments. We propose a Constraint Mechanism (CM) to improve generated format accuracy. Our results show SC accuracy increased by 1.13 points and NER by 1.06 points in SCNM compared to standalone tasks, with CM raising format accuracy from 63.61 to 100. The findings indicate mutual reinforcement effects between SC and NER, and integration enhances both tasks' performance. We additionally implemented the SLG framework on single SC task. It yielded superior accuracies compared to the baseline on two distinct Japanese SC datasets. Notably, in the experiment of few-shot learning, SLG framework shows much better performance than fine-tune method. These empirical findings contribute additional evidence to affirm the efficacy of the SLG framework.

## 1 Introduction

In the realm of information extraction, numerous specialized tasks exist, such as named entity recognition (Ritter et al. 2011; Lample et al. 2016; Nadeau and Sekine 2007), relation extraction (Mintz et al. 2009) event extraction (Etzioni et al. 2008), sentence classification (Zhang et al. 2015), sentiment analysis (Medhat et al. 2014; Pang et al. 2002), and more. With the advent of the Transformer architecture, pre-training and fine-tuning paradigms have gained widespread adoption. Typically, models undergo unsupervised pre-training on a large-scale, general corpus, such as Wikipedia text, in order to acquire foundational knowledge. These pre-trained models are then fine-tuned for specific downstream tasks. However, due to the considerable variation in data features and requirements across tasks, adapting a single dataset and model to multiple tasks simultaneously is challenging. Consequently, researchers often create dedicated datasets for distinct IE tasks and employ these for fine-tuning pre-trained models. Moreover, IE methods have evolved from sequence labeling tasks utilizing Long Short-Term Memory (LSTM) (Huang et al. 2015) to sequence-to-sequence generative IE methods (Lu et al. 2022). The emergence of generative approaches indicates the feasibility of addressing multiple tasks with a single model by unifying input and output formats. In the present study, illustrated by Figure 1, the SC and NER tasks are fed as inputs to their respective fine-tuned models. Then models generate corresponding labels/spans for each task, respectively.

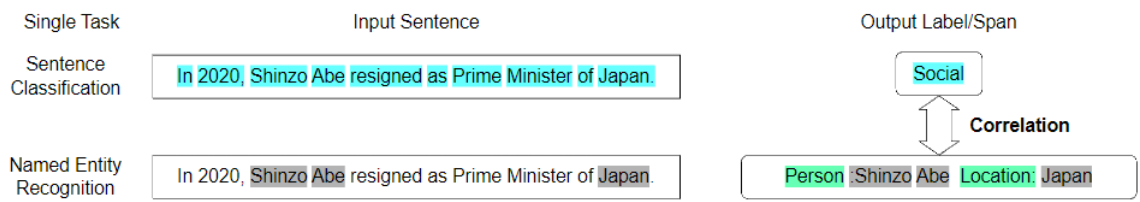

**Figure 1** The illustration depicts the interrelationship between the labels of Named Entity Recognition (NER) and Sentence Classification (SC) within a single sentence.

While generic pre-training knowledge can be beneficial for various downstream tasks, the possibility of mutual reinforcement effects between tasks remains an open question. To explore this, we hypothesize that mutual reinforcement effect exists between different tasks. To test this hypothesis, we focus on Named Entity Recognition (NER) and Sentence Classification (SC) as the most representative tasks in IE, as illustrated in Figure 1. In the SC task, a model generates sentence classification labels given an input sentence. In the NER task, a model identifies entity spans in the input sentence and generates corresponding labels and spans. Many task scenarios require simultaneous sentence classification and entity extraction, but no existing dataset satisfies both requirements. Furthermore, SC and NER tasks exhibit correlations in the labels they extract. For instance, a sentence mentioning "Shinzo Abe" likely pertains to Social, and a social sentence is more likely to contain names of social figures and countries. Consequently, we investigate whether leveraging the interrelationship between SC and NER tasks can improve model performance.

In this context, we propose a novel framework for handling both Japanese SC and NER tasks. The primary contributions of this work include:

1. Integrating SC and NER tasks into a new Sentence Classification and Named Entity Recognition Multi-task (SCNM) task and constructing an SCNM dataset by annotating SC labels on the existing Wikipedia Japanese NER dataset.
2. Proposing a Sentence-to-Label Generation Framework(SLG) for addressing the SCNM task, comprising format converter construction, incremental training, and the development of a format Constraint Mechanism(CM). The format converter enables the SLG to handle the SCNM task, as well as SC or NER tasks separately, highlighting its generalizability.
3. Demonstrating through ablation experiments that SC and NER tasks have mutual reinforcement effects. The performance of a model trained by combining both tasks surpass that of a model fine-tuned on a single task, supporting the notion that 1+1>2. This finding offers insights for future scenarios requiring SCNM.
4. The foundational model of the SLG framework was retrained incrementally utilizing the SCNM dataset. This retrained SLG framework was then applied to various other Japanese SC tasks, resulting in enhanced performance when compared to the tasks completed without the SLG framework. Furthermore, the application of few-shot learning in conjunction with the SLG framework greatly surpassed the performance of the fine-tune method.

## 2 Related Work

In this section, we provide a comprehensive overview of previous studies on generative IE methodologies. Furthermore, we delineate the similarities and distinctions between these prior works and our current research, thereby highlighting the novel contributions of our study.

**Word-level**. In (Yan et al. 2021), the authors propose a novel sequence-to-sequence framework that addresses flat, nested, and discontinuous NER subtasks through entity span sequence generation. Similarly, (Chen et al. 2022) used a self-describing mechanism for few-shot NER, which leverages mention describing and entity generation. GenIE (Josifoski et al. 2022) uses the transformer model to extract unstructured text relationally through global structural constraint. And LightNER (Chen et al. 2022) is addresses class transfer by constructing a unified learnable verbalizer of entity categories and tackles domain transfer with a pluggable guidance module. InstructionNER (Wang et al. 2022) and UIE (Lu et al. 2022) have also developed frameworks for word-level IE tasks.

**Sentence-level**. In terms of using sentences label to improve the NER effect. Joint learning framework used BiLSTM model and attention, CRF layer to improve the effectiveness of NER through sentence labeling (Kruengkrai et al. 2020). MGADE that uses a dual-attention mechanism to concurrently address two Adverse Drug Event (ADE) tasks: ADE entity recognition at the word level (fine-grained) and ADE assertive sentence classification (coarse-grained). The model takes advantage of the interdependencies between these two levels of granularity to improve the performance of both tasks (Wunnava et al. 2020).

In conclusion, this study employed a generative model for Sentence-to-Label conversion, distinguishing itself from previous studies by pioneering the utilization of mutual reinforcement effects between SC and NER tasks. This novel approach effectively enhanced the accuracy of both tasks. Additionally, we introduced the SLG framework, which enables the model to adeptly manage both SC and NER tasks simultaneously.

## 3 SCNM Task Setup and Dataset Constructed

Before delving into the SLG, let us first describe the structure of the SCNM task. SCNM involves the classification of both sentence-level and word-level information within a single sentence. As illustration of Figure 2, given an input sentence (i.e., In 2020, Shinzo Abe resigned as Prime Minister of Japan), the model is expected to generate a classification label (i.e., Social) for the sentence, along with named entity labels (i.e., Person, Location) and the corresponding word spans (i.e., Shinzo Abe, Japan) present in the sentence. Moreover, when selecting and constructing the dataset, it is crucial to encompass a wide range of content, and the classification labels should be generic rather than specific to a narrow domain. Considering these factors, we chose the Japanese Wikipedia-based NER dataset (Omiya 2021) for our SCNM dataset foundation. It consists of 5,343 sentences (4,859 with named entities, 484 without) and includes 8 categories (13,185 total named entities): person, company, political org., other org., location, public facility, product, and event. This diverse and broad dataset is ideal for the SCNM task.

Following the selection of the NER dataset, we annotated sentence classification labels based on the original dataset. We partitioned the Wikipedia sentences into five primary categories: social, literature and art, academic, technical, and natural. All 5,343 sentences were categorized into these five groups, ultimately resulting in the construction of the SCNM dataset. Figure 2 illustrates a specific instance of the input and output by the SCNM task.

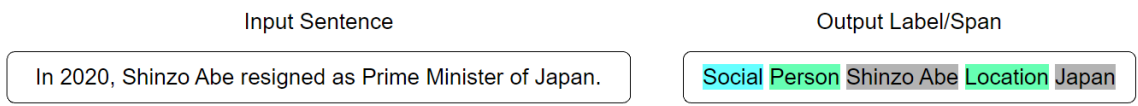

Figure 2 The illustration of Sentence Classification and Named Entity Recognition Multi-task (SCNM).

## 4 Evaluation

Upon constructing the SCNM dataset, the primary challenge we encounter is devising a comprehensive set of evaluation metrics for this new dataset. We categorize the SCNM evaluation metrics into two distinct classes: text evaluation and format evaluation.

**Text Evaluation.** As illustrated in Figure 3, we present several instances of both generated and actual text, labeled numerically (1-5) for further analysis. Given that the combined task of SC and NER precludes the computation of traditional metrics (e.g., precision recall f1-score) in the conventional manner, generative IE deviates from the traditional sequence tagging model. Specifically, in the generative IE approach, a token does not correspond to a label. Furthermore, the number of generated named entities and word spans is variable (e.g., 3), as is the number of generated texts, which may also contain extraneous text (e.g., 4 in Figure 3).

**Figure 3** The illustration of text and format evaluation. 1 and 0 represents a correctly or incorrectly generated text, and the total number of all correctly generated text in the test set adds up to $C$.

Consequently, we employ accuracy as the evaluation metric, when generate text 1 is all matched with actual text 5, it counts as one correct text. Although this evaluation criterion may seem somewhat stringent, it is well-suited for SCNM tasks. We denote the set of generated texts as $G$ and the set of actual texts as $A$. The $|A \cap G|$ is represented by $C_{\text{generated text}}$ (the total number of matches texts between $G$ and $A$), while the total number of actual text (i.e., total number of samples in the test set) is denoted by $T_{\text{actual text}}$. In addition, each generated text and the actual text correspond sequentially. So, there is no case where a wrongly generated text is the same as the other actual texts, which results in incorrect calculation of the correct number.

$$SCNM\ Accuracy = \frac{C_{\text{generated text}}}{T_{\text{actual text}}} \tag{1}$$

In addition, we also computed accuracy for SC and NER in the SCNM task separately. Specifically, the SC-label and all remaining NER-label/span in generated text and actual text were split. Then the accuracy is calculated separately.

$$SC\ Accuracy = \frac{C_{SC}}{T_{SC}} \tag{2}$$

$$NER\ Accuracy = \frac{C_{NER}}{T_{NER}} \tag{3}$$

**Format Evaluation.** In generative IE, a key metric is the ability to produce outputs in the right format. Due to the uncontrollable nature of generative models, there's a high chance of generating duplicate words or incoherent texts. Incorrect output formats likely lead to wrong label classification (e.g., 4 in Figure 3). Thus, we augment text evaluation with format generation accuracy assessment.

This additional evaluation aims to gauge a model's proficiency in controlling the generated format. If the first generated text becomes SC-label and the subsequent generated ones are NER-label and span, it is counted as one correct format correct (regardless of whether the generated SC-label and NER-label are correct or not, they are counted as format correct). As illustrated in Figure 3 (e.g., 1 2 3 in Figure 3). The total number of generated texts with correct format is $C_{format}$, and the total number of actual texts is $T_{format}$. The Format Accuracy is defined as:

$$Format\ Accuracy = \frac{C_{format}}{T_{format}} \qquad (4)$$

## 5 Sentence-to-Label Generation Framework

In the preceding section, the SCNM task setting, and dataset construction were presented. This section offers a thorough overview of the Sentence-to-Label Generation (SLG) framework, followed by a detailed explanation of each component of the SLG in three separate subsections.

An overview of the SLG framework is depicted in Figure 4. Initially, the Shinra NER corpus[1] is reformatted using a format converter. Subsequently, the transformed corpus serves as a basis for incremental learning(fine-tune) in the model. The SCNM dataset, converted by the format converter, is then input to the model*. Because the incremental learning Shinra NER corpus is not the same as the content of the SCNM data. So, the format of the format converter used for incremental learning is also different with SLG framework (The specific format will be described in sections 5.1 and 5.2, respectively). Prior to the model's Decoder generating prediction results, a Constraint Mechanism (CM) is incorporated to enhance the model's format generation capabilities. Lastly, the SC-label, NER-label, and corresponding word span are sequentially outputted.

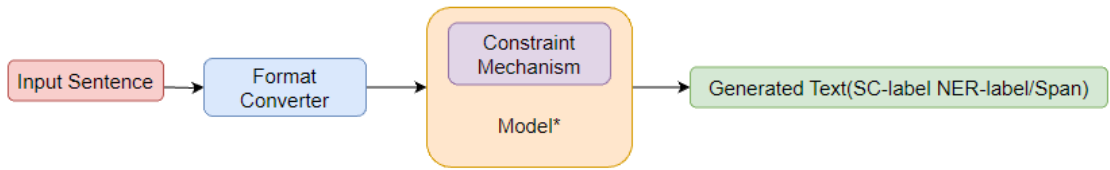

**Figure 4**　The illustration of overview for Sentence-to-Label Generation (SLG) Framework. The model* is the vanilla model with incremental learned using Shinra NER corpus.

---

[1] http://shinra-project.info/shinra2020jp/data_download/

## 5.1 Format Converter

In this subsection, we explore multiple converter formats and evaluate performance in later experiment sections. The most effective format is chosen for our converter.

As show in Figure 5, the optimal format is determined from the experimental outcomes. For the input sequence of the model, the original sentence is positioned at the beginning, succeeded by five SC-label words. The start and end tokens, "<" and ">", respectively, enclose all SC-labels. Subsequently, the original sentence is repeated immediately after the above end mark token ">". To signal the model's initiation of NER-label generation and corresponding word span, a prompt word "NER" is appended after the sentence.

Due to the presence of negative sentences in the SCNM dataset that lack named entities, an additional "None" label is introduced for negative cases, augmenting the original eight NER-labels. Consequently, a total of nine NER-labeled words follows the prompt word. To indicate the commencement and termination of the NER-label word, the start and end tokens, ":", and ";", are employed, respectively. The distinct mark tokens for SC and NER labels demonstrate superior performance in the experiments, as compared to identical mark tokens.

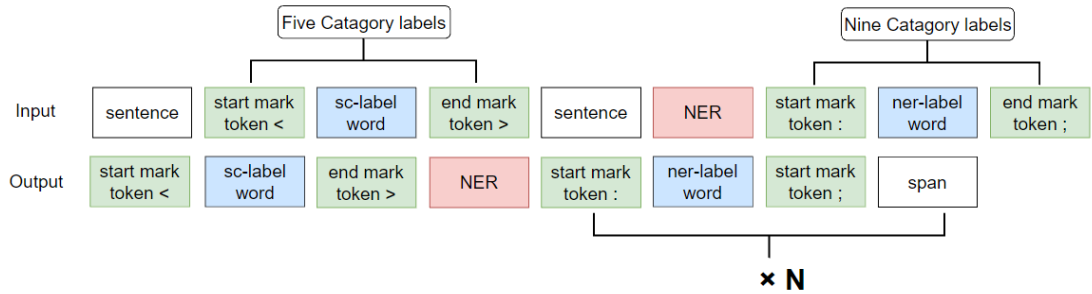

**Figure 5** The illustration of format converter.

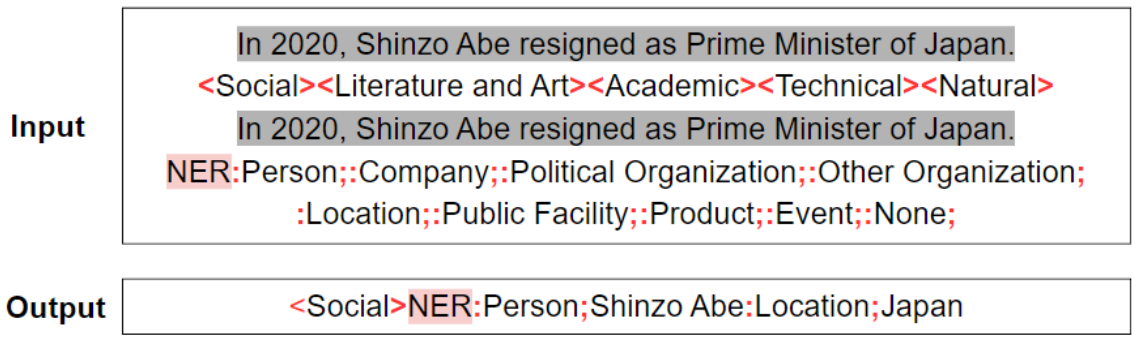

**Figure 6** A specific example of an input converted using the format converter, and the corresponding generated output.

Regarding the output format of the model, the overall and input token order is maintained consistently. However, it is crucial to note that only one of the predicted SC-label words is utilized, rather than all five words present in the input. Following the starting word "NER", the NER-label word and corresponding word span derived from the sentence are provided. Utilizing the format converter, the model demonstrates versatility in managing diverse tasks, such as SC, NER, and SCNM. These tasks can be effectively addressed either individually or concurrently. And through format converter, the model learns the correct format to generate a uniform format. A specific example of the input and output sequence is show in Figure 6.

### 5.2 Incremental Learning

In this subsection, we elucidate the process of applying incremental learning (IL) to the vanilla T5[2] and BART[3] model using the Shinra NER corpus, as well as the underlying rationale. The model is primarily pre-trained for sequence-to-sequence tasks in text processing (Raffel et al. 2020). However, it lacks specific pre-training for word-level attributes, such as named entities. Our objective is to implement IL in a sequence-to-sequence format, tailored for named entities, without causing the model to lose its pre-trained knowledge.

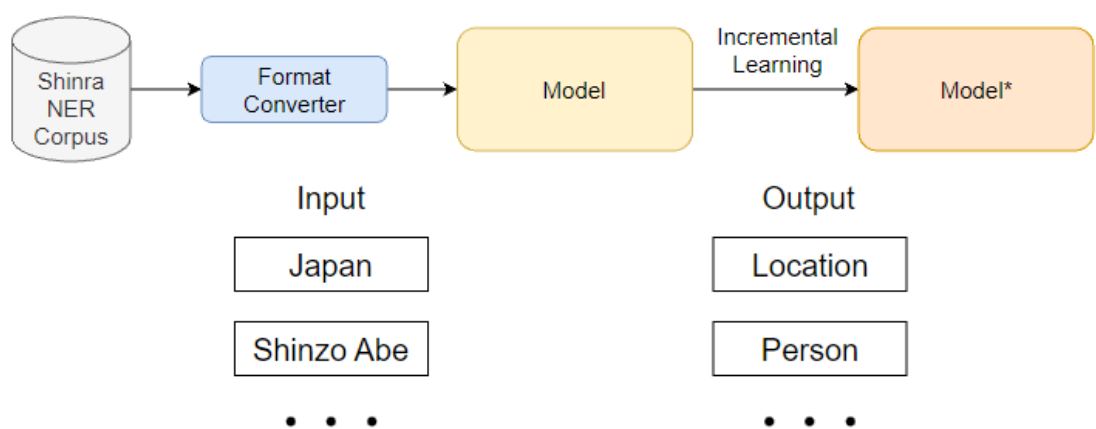

**Figure 7**  A specific example of an input converted using the format converter, and the corresponding generated output of Shinra NER corpus.

We selected the Shinra2020-JP corpus as the data source for IL. The Shinra project, an extended named entity recognition (NER) endeavor, is constructed based on the Japanese Wikipedia. Consequently, the corpus encompasses a wide array of named entities derived from Wikipedia. In other words, this corpus contains various named entities of Wikipedia. The categories of named entities are facility, event, organization, location, airport name, city, company, Compound, person. It is a relatively comprehensive and extensive NER corpus. The dataset is used for incremental learning of the model after a simple format transformation (e.g., input: Japan, output: Location). The total number of samples is 14117 and there are no duplicate samples. As illustrated in Figure 7, the model employed within the SLG framework is ultimately acquired.

## 5.3 Constraint Mechanism

To enhance the accuracy of the model's output format, we introduce an efficient and straightforward Constraint Mechanism (CM). Let $X_1$ denote the initial token predicted by the Decoder, with the total predicted sequence comprising $n$ tokens. The predicted sequence can be represented as $(X_1, X_2, ..., X_n)$. The prediction probability for the second token $X_2$ can be formulated as the subsequent equation:

$$P(X_2|X_1) = Decoder_2(X_1, Encoder(Inputs)) \qquad (5)$$

Here, $Encoder(Inputs)$ denotes the vector resulting from the computation of the input by the next layer of Encoder. $Decoder_2$ denotes the result of the computation of the vector output by encoder and the first token vector output by Decoder of the first layer is passed to the second layer Decoder.

In Figure 8, a specific example is presented. The output text refers to the desired output sequence that should be generated accurately. Notably, the first token of every output text within the SCNM dataset remains constant, represented by the "<" symbol. Consequently, the initial token of each predicted sequence is compelled to be replaced with the "<" token. In other words, this substitution corresponds to the numerical value of the "<" symbol in the model's vocabulary.

Adopting this approach ensures that the first predicted token in each newly generated output text is accurate, which in turn enhances the precision of subsequent token predictions. The ultimate experimental outcomes corroborate that the CM technique effectively augments the model's capacity to generate accurate formats, thereby improving the overall correctness of SCNM tasks.

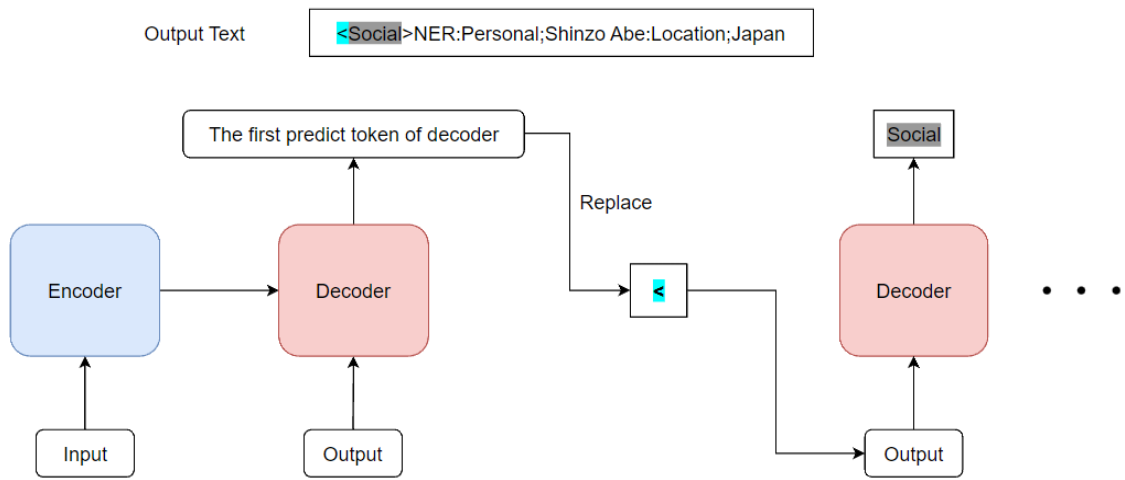

**Figure 8** The processes of Constraint Mechanism (CM).

# 6 Implement SLG framework to individual SC tasks

We also apply the SLG framework to a separate SC task, to test the performance of the SLG framework. As shown in Figure 9, we incrementally trained the original T5 model with the Shinra NER corpus. The SCNM dataset is then used for incremental training. The obtained T5 model is used in the SLG framework.

In Figure 10, we present a classification task aimed at discerning the relationship between two sentences. The labels used in this task are 'entailment', 'neutral', and 'contradiction'.

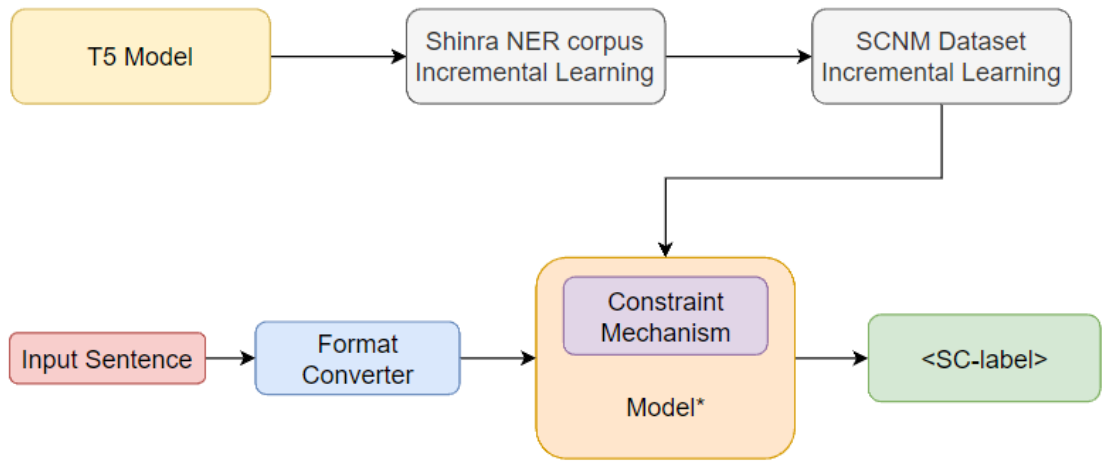

**Figure 9**   The illustration of overview for Sentence-to-Label Generation (SLG) Framework applies in individual SC tasks. Model* is a T5 model trained in two incremental learning.

To facilitate this task, we manipulate the original dataset with the aid of a format converter. Each input sentence is appended with the three labels, designated as SC-labels, that are bracketed by start and end mark tokens. Notably, the predicted SC-label, which is a labeled word encompassed by mark tokens, is also appended. This simple modification enables us to apply the SLG framework to an isolated SC task. Thus, we demonstrate the adaptability of the SLG framework in accommodating diverse linguistic tasks.

Although the SCNM dataset is not related to the two sentence relation classification tasks. We nevertheless added the SCNM dataset to the incremental training. We want the model to learn the ability to generate the format correctly.

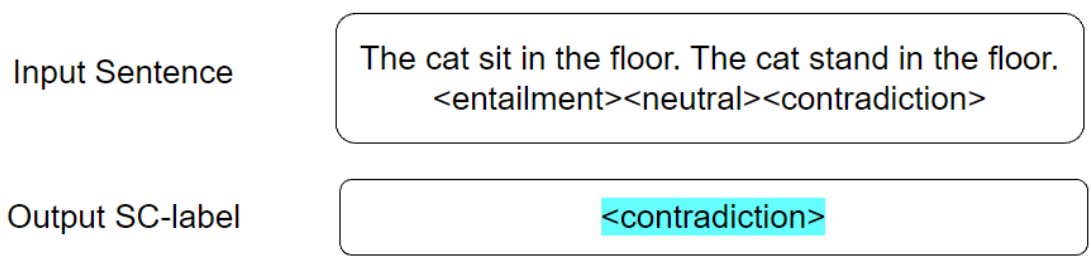

Figure 10 A specific example about the SC tasks transformed by Format Converter

## 7 Experiments

In this chapter, we present a comprehensive series of experiments conducted on our proposed SLG framework and its application to SCNM tasks. Our aim is to demonstrate the effectiveness of the SLG framework and the synergistic effect of integrating the SC and NER tasks.

Given that the Japanese T5 model is only available in base size, we employ the T5-base as the underlying model weight for the SLG framework. In addition to our initial methodology, we integrated the BART base and large model into our study. This model, renowned for its encoder-decoder architecture, was employed as the foundational model within the SLG framework. This integration was executed with the primary objective of investigating the reciprocal reinforcement effect between SC and NER tasks. To ensure robustness, we conduct each experiment three times, and the final result is obtained by averaging the outcomes. The proportion used for dividing the data into the training set and the test set is 9:1. We adopt a randomized approach to data set partitioning, using the parameter "Random Seed = None" to guarantee distinct training and testing sets for each iteration. To mirror the conditions of all experiments more accurately, we established the random seed of the dataset partition to be 123123. This was done to ensure the consistency of the training and testing sets for each iteration. We also re-run all experiments (including ablation experiments) in this way. For an in-depth discussion of the evaluation metrics utilized, please refer to Section 3.1.

Due to the significant variability between the T5-base Japanese model and the BART-base/large Japanese model, we have segmented our analysis into two distinct chapters. Each chapter is dedicated to a comprehensive examination of the experimental results generated by each model, providing an in-depth interpretation and discussion separately.

Finally, we leveraged two sentence classification tasks within the Japanese General Language Understanding Evaluation (JGLUE) (Kurihara et al. 2022) to assess the efficacy of the SLG framework on individual SC tasks. This experimental setup was orchestrated to investigate whether the SLG framework augments the performance of various individual SC tasks as well.

Additionally, we initiated a few-shot learning experiment involving two specific Japanese sentence classification datasets. A comparative analysis was performed on the model's performance in both 5-shot and 10-shot scenarios, with and without the incorporation of the SLG framework, thereby providing valuable insights into its potential benefits and applicability.

The statistical information of the three data sets is shown in Table 1. The "→" representatives were converted from 26 categorical tasks to 6 categorical tasks.

Table 1　The statistical information of experiment dataset.

| Dataset | Number of Samples | Number of Categories |
|---|---|---|
| SCNM | 5343 | 5 SC-label & 10 NER-label |
| JNLI | 20073 | 3 |
| JSTS | 12451 | 26 → 6 |

## 7.1 Compare the effect of different formats on results

Before embarking on further experiments, it is crucial to determine an optimal unified format. To achieve this, we designed five distinct formats and selected the most effective one as the conversion format, informed by the empirical results of our tests. In this context, we exclusively utilized the T5 model, given its superior performance compared to the BART model. Table 2 displays these formats, where the first row represents the input text, and the second row signifies the output text. Given that the number of NER-label span pairs generated in each sentence is indefinite, we use (...)*x to denote the count of generated NER-label span pairs. The "sentence"[1] represents the sentence that originally needed to be categorized and entities extracted. Moreover, since there are five SC-labels in the input text, we represent this with *5. The corresponding NER-labels total nine, denoted by *9, which includes eight labels from the original dataset and one additional "None" label.

In the experimental setup, we aimed to minimize the influence of external factors on the results. Therefore, only the original T5-base model was employed in the format comparison experiment, with the random seed and other hyperparameters held constant.

Table 2 reveals that the accuracy of the simplest format is considerably low on the SCNM dataset, particularly when the input consists solely of sentences. In this case, the accuracy is 0, with all results being incorrect. This is due to the model's inability to generate the desired format accurately. As a result, even if the SC or NER labels are generated correctly, they are still considered errors based on the strict evaluation criteria. The second format introduced a simple prompt word, "sentence NER", which slightly improved accuracy to 0.19; however,

---

[1] The "{ }" included "sentence" is added to emphasize the representation. It is not in the actual dataset.

the accuracy remained substantially low.

Table 2    The result of compared different format with SCNM tasks. And without IL and CM. The first line is input text, and second line is output text.

| Different Format | Accuracy |
|---|---|
| {sentence}<br>:SC-label;(:NER-label;span)*x | 0 |
| Sentence:{sentence}<br>label:SC-label;NER(:NER-label;span)*x | 0.19 |
| {sentence}category(:SC-label;)*5{sentence}NER(:NER-label;)*9<br>category:SC-label;NER(:NER-label;span)*x | 0.37 |
| {sentence}(:SC-label;)*5{sentence}NER(:NER-label;)*9<br>:SC-label;NER(:NER-label;span)*x | 0.56 |
| {sentence}(<SC-label>)*5{sentence}NER(:NER-label;)*9<br><SC-label>NER(:NER-label;span)*x | 29.40 |

In the third and fourth formats, we incorporated all the SC-labels (five label words) and NER-labels (nine label words) into the input text. Upon adding sufficient prompt words to the input text, the accuracy of the third format increased to 0.37 and 0.56 respectively. Lastly, we modified the start and end mark tokens of the SC-label to "<" and ">", respectively, to facilitate the model's differentiation between SC-label and NER-label. Consequently, the accuracy significantly improved to 29.40, surpassing the results of all previous formats. This format comparison experiment highlights the critical role of prompt design in obtaining accurate outcomes.

## 7.2 Result on T5 Model

### 7.2.1 Result on SLG with SCNM task

In this subsection, we conduct a comprehensive evaluation experiment utilizing the SLG framework. Table 3 presents the results, with the first row displaying the metrics SCNM, SC, and NER, which correspond to the overall accuracy of the SCNM dataset, the accuracy of the SC dataset individually, and the accuracy of the NER dataset individually.

Three distinct datasets are outlined in the first column. "SCNM*" represents the full SCNM dataset, encompassing both SC and NER tasks. "SC Only" and "NER Only" signify the evaluation of the SC and NER datasets independently. To achieve this, we employ a format converter to separate the SC and NER components within the SCNM dataset, resulting in two distinct datasets (i.e., SC dataset and NER dataset). All three datasets are assessed using the SLG framework.

Table 3    The result of SLG framework with SCNM dataset. Accuracy was used for all evaluation metrics.

| Dataset | SCNM Acc. | SC Acc. | NER Acc. | Format Acc. |
|---|---|---|---|---|
| SCNM | **72.41** | **88.89** | **81.96** | **100** |
| SC Only | − | 87.76 | − | − |
| NER Only | − | − | 80.90 | − |

The evaluation reveals that the SCNM task attains a notable score of 72.41, even under rigorous evaluation metrics. Benefiting from Constraint Mechanism, the format accuracy reaches 100. Furthermore, by integrating the SC and NER tasks, the accuracy of NER improves by 1.06 compared to evaluating the NER task individually. Similarly, the SC task accuracy increased by 1.13 compared to evaluating it separately. The experimental outcomes highlight the exceptional performance of the SLG framework in handling the SCNM task, as well as the mutual reinforcement effects observed between the SC and NER tasks.

### 7.2.2 Ablation Experiment

To assess the impact of SL and CM on the outcomes within the SLG framework, we carried out a series of ablation experiments. Table 4 illustrates that, upon removing Incremental Learning, the NER accuracy within the SCNM dataset experienced a substantial decline of 15.98. Concurrently, the SC in the SCNM dataset also saw a reduction of 3.18. These findings highlight that incremental training not only enhances NER in the SCNM dataset but also exerts a positive influence on SC. This observation aligns with our initial hypothesis that SC and NER tasks exhibit a mutual reinforcement effect.

Conversely, when CM was eliminated from the SLG framework, the format accuracy of the model plummeted to 63.61. Consequently, the SCNM, SC, and NER metrics also witnessed significant decreases, falling to 46.38, 55.99, and 52.62, respectively. These experimental outcomes underscore the efficacy of our proposed CM approach. By effectively managing the initial token generated by the model, we can guide the subsequent tokens towards accurate generation, thereby improving the model's overall performance.

Table 4    The result of IL or CM with ablation experiment.

| Method | SCNM Acc. | SC Acc. | NER Acc. | Format Acc. |
|---|---|---|---|---|
| SLG | **72.41** | **88.89** | **81.96** | **100** |
| w/o IL | 56.18 | 85.71 | 65.98 | 99.50 |
| w/o CM | 46.38 | 55.99 | 52.62 | 63.61 |
| w/o IL & CM | 3.10 | 5.49 | 4.68 | 6.30 |

Finally, when IL and CM were removed, all four accuracies dropped to a very low level. This also reflects the importance of IL and CM in the SLG framework. Despite not being explicitly mentioned in the table, the absence of a Format Converter would exacerbate this result, reducing it even further to 0. Ablation experiments also demonstrate the indispensability and effectiveness of the three components of the SLG framework, like the results in Subsection 5.1.

## 7.3 Result on BART Model

In this chapter, we will use the BART base/large model as the base model of the SLG framework. To test the performance of the SLG framework.

### 7.3.1 Result on SLG with SCNM task

As indicated in Tables 5 and 6, the accuracy of the NER significantly decreases upon employing the BART model, yielding NER accuracy rates of 33.15 and 32.71 respectively. In turn, this decrease in NER accuracy precipitates a substantial reduction in the corresponding SCNM accuracy, which plummets to 27.40 and 27.22 respectively. Notably, the results observed for the base and large configurations of the model do not demonstrate substantial variance.

The disparate performances between the BART and T5 models can be attributed to various factors, chiefly among which are the architectural complexities and the difference in the size and quality of the pre-training corpuses utilized. The BART model is inherently less complex than the T5 model and demonstrates a higher susceptibility to overfitting.

Additionally, a significant divergence in their training datasets impacts the performance of these models. The pre-training corpus for the T5 model comprises three distinctive data sources: Wikipedia, OSCAR, and CC-100, collectively forming a substantial pool of approximately 100GB of Japanese corpus. On the other hand, the BART model's pre-training is confined to a relatively meager 3GB Wikipedia corpus. This contrast in the size and richness of the training corpuses has observable ramifications on the performance. Empirical results suggest that a more extensive and diverse pre-training corpus can markedly enhance the model's competency in entity recognition and entity generation.

It is noteworthy that the BART model does not achieve perfect accuracy in formatting, falling slightly short at 99.94. An analysis of the generated results unveils some anomalies. For instance, one of the outcomes is "<Company;nabisuko". Despite our efforts to mandate the initial token generation as "<", the subsequent model output does not accurately reproduce the SC-label. Instead, it produces the NER-label directly. This discrepancy can be partially attributed to the overlearning issues inherent to the BART model.

Table 5  The result of BART-base model with SLG framework.

| Dataset | SCNM Acc. | SC Acc. | NER Acc. | Format Acc. |
| SCNM | <u>27.40</u> | 84.77 | <u>33.15</u> | <u>99.94</u> |
| SC Only | — | 83.65 | — | — |
| NER Only | — | — | <u>25.59</u> | — |

Table 6  The result of BART-large model with SLG framework.

| Dataset | SCNM Acc. | SC Acc. | NER Acc. | Format Acc. |
|---|---|---|---|---|
| SCNM* | <u>27.22</u> | 85.58 | <u>32.71</u> | 100 |
| SC Only | — | 80.34 | — | — |
| NER Only | — | — | <u>28.71</u> | — |

### 7.3.2 Ablation Experiment

We also conducted a comprehensive ablation study on the BART-base/large model. As illustrated in Tables 7 and 8, the individual results display negligible disparities. In the case of the BART-base model, after the IL component was removed, there was a noticeable decrease in NER accuracy. Conversely, SCNM and SC accuracies exhibited an increase.

Contrastingly, in the BART-large model, both SCNM and SC accuracies reached peak performance, while NER accuracy saw an enhancement after the removal of the CM. The apparent reason behind these findings is speculated to be linked to the overfitting issue intrinsic to the BART model. During the fine-tuning process utilizing various methodologies, overfitting ensued, which led to the slight variation in the ultimate results.

Table 7  The result of IL or CM with ablation experiment in BART-base Model.

| Method | SCNM Acc. | SC Acc. | NER Acc. | Format Acc. |
|---|---|---|---|---|
| SLG | 27.40 | 84.77 | 33.15 | 99.94 |
| w/o IL | <u>27.90</u> | <u>85.27</u> | 32.83 | <u>100</u> |
| w/o CM | 27.40 | 84.71 | <u>33.33</u> | 100 |
| w/o IL & CM | 25.41 | 84.46 | 30.77 | 100 |

Table 8  The result of IL or CM with ablation experiment in BART-large Model.

| Method | SCNM Acc. | SC Acc. | NER Acc. | Format Acc. |
|---|---|---|---|---|
| SLG | <u>27.22</u> | <u>85.58</u> | 32.71 | <u>100</u> |
| w/o IL | 25.03 | 84.27 | 32.27 | 99.75 |
| w/o CM | 27.03 | 83.77 | <u>33.27</u> | 98.13 |
| w/o IL & CM | 24.34 | 84.90 | 29.71 | 99.69 |

### 7.4 Result on Fixed Random Seed

Due to the stochastic partitioning of the dataset, the end-result of the experiment consistently demonstrated variability. Our objective was to further regulate the independent variables to validate the mutual reinforcement effect. Consequently, we standardized the random seed to 123123, a move that guarantees identical replication of the training and test sets during each iteration. This strategic decision precipitated a comprehensive re-evaluation of our experimental findings.

### 7.4.1 Result on SLG with SCNM task

Table 9. 10. 11 distinctly highlight the significant disparities in results upon utilizing an identical random seed. Particularly noteworthy is the impact on the T5 model when NER and SC tasks are separated; the NER accuracy plummets to zero. This abrupt reduction can be attributed to the model's inability to accurately generate a comprehensive NER-label and span.

Parallel observations can be made from the results of the BART-base/large models, where a substantial decrease in accuracy is evident after segregating the SC and NER tasks. This pattern underscores a mutual reinforcement effect between the SC and NER tasks.

Table 9　The result of T5-base model with SLG framework.

| Dataset | SCNM Acc. | SC Acc. | NER Acc. | Format Acc. |
|---|---|---|---|---|
| SCNM* | 73.41 | 91.20 | 80.71 | 100 |
| SC Only | − | 87.83 | − | − |
| NER Only | − | − | 0 | − |

Table 10　The result of BART-base model with SLG framework.

| Dataset | SCNM Acc. | SC Acc. | NER Acc. | Format Acc. |
|---|---|---|---|---|
| SCNM* | 26.78 | 86.70 | 31.46 | 100 |
| SC Only | − | 85.39 | − | − |
| NER Only | − | − | 26.22 | − |

Table 11　The result of BART-large model with SLG framework.

| Dataset | SCNM Acc. | SC Acc. | NER Acc. | Format Acc. |
|---|---|---|---|---|
| SCNM* | 28.28 | 85.02 | 33.33 | 100 |
| SC Only | − | 83.90 | − | − |
| NER Only | − | − | 29.03 | − |

### 7.4.2 Ablation Experiment

In this section, we fixed random seed and conduct ablation experiments on the three base models of the SLG framework. As shown in Tables 12, 13 and 14, although on the BART-base/large model, the difference of each result is not significant. But on the T5-base model, the accuracy gradually decreases with the removal of IL and CM. The effectiveness of IL and CM in SLG framework was proved.

Table 12  The result of IL or CM with ablation experiment in T5-base Model.

| Method | SCNM Acc. | SC Acc. | NER Acc. | Format Acc. |
|---|---|---|---|---|
| SLG | <u>73.41</u> | <u>91.20</u> | <u>80.71</u> | <u>100</u> |
| w/o IL | 58.24 | 87.27 | 67.23 | 100 |
| w/o CM | 51.12 | 65.17 | 57.30 | 72.28 |
| w/o IL & CM | 35.39 | 53.37 | 40.64 | 59.55 |

Table 13  The result of IL or CM with ablation experiment in BART-base Model.

| Method | SCNM Acc. | SC Acc. | NER Acc. | Format Acc. |
|---|---|---|---|---|
| SLG | 26.78 | <u>86.70</u> | 31.46 | 100 |
| w/o IL | <u>27.90</u> | 85.77 | <u>32.40</u> | 100 |
| w/o CM | 26.78 | <u>86.70</u> | 31.46 | 100 |
| w/o IL & CM | 27.90 | 85.77 | <u>32.40</u> | 100 |

Table 14  The result of IL or CM with ablation experiment in BART-large Model.

| Method | SCNM Acc. | SC Acc. | NER Acc. | Format Acc. |
|---|---|---|---|---|
| SLG | <u>28.28</u> | 85.02 | 33.33 | <u>100</u> |
| w/o IL | 24.34 | <u>86.33</u> | 27.90 | 99.63 |
| w/o CM | <u>28.28</u> | 82.21 | <u>33.52</u> | 96.63 |
| w/o IL & CM | 24.34 | <u>86.33</u> | 27.90 | 99.63 |

### 7.5 Result on Individual SC tasks

In this subsection, we aim to experimentally evaluate a classification task involving two distinct pairs of sentences from the JGULE dataset. We will analyze the performance of these tasks, contrasting the results with and without the implementation of the SLG framework.

### 7.5.1 Result of Individual SC tasks and Ablation Experiment

The primary focus of both tasks is to classify the relationship between two sentences. However, the tasks diverge in terms of their label configurations. The first sentence pair task operates on a triple classification dataset(JNLI), utilizing the labels 'entailment', 'neutral', and 'contradiction'. The second sentence pair classification task(JSTS), on the other hand, assigns a score between 0 and 5 at intervals of 0.2. In an effort to streamline the classification process, we have reduced the original 26 labels down to just six. This simplification involved adjusting the intervals to '1' and consolidating the scoring range to whole numbers between 0 and 5. This approach refines the task into a six-category sentence classification assignment.

As illustrated in Table 15, the outcomes of applying the SLG framework to the tri-classification JNLI dataset are detailed. We have chosen to present the results of the three distinct experiments individually, with the aim of highlighting the stability and robustness of the SLG framework. By delineating the results of each experiment, we can better illustrate the variability within each outcome.

Initially, with the incorporation of the SLG framework, the mean accuracy peaks at 90.76. Conversely, upon the exclusion of the SLG framework and relying solely on the simple fine-tune methodology, the mean accuracy undergoes a decline, reaching 84.21. The simple fine-tune method here is after removing FC,CM and IL. The result of simple fine-tune on the original T5 model using the original SCNM dataset without using the SLG framework at all. These paired comparative analyses emphatically underscore the instrumental role of the SLG framework in augmenting the efficacy of a singular SC classification task.

Table 15　The result of SLG framework in JNLI dataset.(3 times result and Average Acc.)

| Method | Accuracy 1 | Accuracy 2 | Accuracy 3 | Average Acc. |
|---|---|---|---|---|
| SLG | **91.88** | 90.23 | 90.18 | 90.76 |
| w/o IL | 89.69 | 91.53 | 91.28 | **90.83** |
| w/o CM | <u>67.26</u> | **91.88** | **92.48** | 83.87 |
| w/o FC & CM | 83.51 | 85.45 | 83.01 | 83.99 |
| Simple Fine-tune (w/o FC, IL & CM) | 83.06 | 83.91 | 85.65 | 84.21 |

Table 16 demonstrates the experimental outcomes for the six-category JSTS dataset. It's evident that an increased level of difficulty corresponds to a significant decline in all results. However, the average accuracy associated with the SLG framework, standing at 57.48, considerably outperforms the fine-tuning method's 35.18. Moreover, the SLG framework maintains a stable accuracy level, showcasing its robustness. On the contrary, substantial fluctuations are observed in the accuracies achieved via the fine-tuning method.

The ablation experiment yielded results parallel to those obtained from the previous dataset, wherein the impact of IL was not pronounced. Interestingly, the combined influence of Format Converter (FC) and CM was considerable. Upon removing FC and CM, the average accuracy plummeted to 42.57. Despite this, the accuracy fluctuation across three instances was not as severe as when the SLG framework was absent.

This implies that IL still has a significant effect. Following the implementation of IL, the model exhibits enhanced stability in generating a standardized format. Consequently, the results are not only consistent but also superior to the fine-tuning method. In conclusion, the incremental learning of the SCNM dataset is also facilitated for the downstream separate SC task.

Table 16　The result of SLG framework in JSTS dataset.(3 times result and Average Acc.)

| Method | Accuracy 1 | Accuracy 2 | Accuracy 3 | Average Acc. |
|---|---|---|---|---|
| SLG | **59.92** | 55.58 | 56.95 | **57.48** |
| w/o IL | 56.55 | **56.39** | **58.47** | 57.14 |
| w/o CM | 53.98 | 54.46 | 52.29 | 53.58 |
| w/o FC & CM | 43.21 | 40.96 | 43.53 | 42.57 |
| Simple Fine-tune (w/o FC, IL & CM) | 39.92 | 22.65 | 42.97 | 35.18 |

### 7.5.1 Result on Few-shot Experiments and Analysis

　　The ablation experiment yielded results parallel to those obtained from the previous dataset, wherein the impact of IL was not pronounced. Interestingly, the combined influence of FC and CM was considerable. Upon removing FC and CM, the average accuracy plummeted to 42.57. Despite this, the accuracy fluctuation across three instances was not as severe as when the SLG framework was absent.

　In an endeavor to further evaluate the efficacy of the SLG framework on a distinct SC task, we extended our study to include experiments on few-shot learning. Specifically, we conducted two sets of experiments, employing both 5-shot and 10-shot learning scenarios. Subsequent to each of these setups, comprehensive fine-tuning of the model was undertaken.

　As can be seen in Table 17, the accuracy of the model maintained considerable stability post the application of the SLG framework. The 5-shot experiments culminated in an average accuracy of 54.78, while the 10-shot learning experiments yielded an accuracy of 55.39. These results underline the SLG framework's superiority in facilitating small-sample learning, as evidenced through the successful implementation of Constraint Mechanism (CM) and Incremental Learning (IL).

Table 17　The few-shot result in JNLI dataset.(3 times result and Average Acc.)

| Method | Accuracy 1 | Accuracy 2 | Accuracy 3 | Average Acc. |
|---|---|---|---|---|
| SLG 5-shot | **54.91** | **54.21** | **55.21** | **54.78** |
| SLG 10-shot | **55.71** | **55.95** | **54.51** | **55.39** |
| Simple Fine-tune 5-shot | 0 | 0 | 0 | 0 |
| Simple Fine-tune 10-shot | 0 | 0 | 0 | 0 |

In stark contrast, the results derived from the use of the simple fine-tuning method were negligible, scoring zero across the board. As depicted in Figure 11, the generated outputs from the few-shot experiments are displayed. The first line is the original Japanese text generated and the second line is the translated English. The left side of the figure presents the results generated by the T5 model, juxtaposed against the actual labels on the right. Evidently, the model was incapable of learning from a minimal sample size to generate the correct labels, producing nonsensical sentences instead.

**Generated Texts** / **Actual Texts**

,山の上に牛が2頭います。。
There are two cows on the mountain.
中立
neutral

,マフラーをした女性が店の前でドーナッツを食べています。,。,
,A woman wearing a scarf is eating donuts in front of the shop. ,. ,
中立
neutral

,りんごがボウルに、りんごがボウルに、。
Then the apples are in a bowl and the apples are in a bowl.
中立
neutral

,男性が2名、女性が1名、女性が1名、。
Two men, one woman, one woman.
中立
neutral

芝生の間の道には車とモーターボートがあります。,。,
There are cars and motorboats on the road between the lawns. ,. ,
中立
neutral

**Figure 11** Example of generated text with simple fine-tune method in few-shot experiment.

**Generated Texts , Actual Texts**

＜矛盾＞, ＜矛盾＞
<contradiction>, <contradiction>

＜矛盾＞, ＜中立＞
<contradiction>, <neutral>

＜中立＞, ＜中立＞
<neutral>, <neutral>

＜中立＞, ＜中立＞
<neutral>, <neutral>

＜中立＞, ＜同義＞
<neutral>, <entailment>

**Figure 12** Example of generated text with SLG framework in few-shot experiment.

Furthermore, as illustrated in Figure 12, the labels generated post the application of the SLG framework were predominantly repetitive. However, it is noteworthy that the model demonstrated the ability to generate outputs in adherence to the pre-established format accurately. This key improvement is considered the principal factor contributing to the enhanced accuracy scores.

Table 18  The few-shot result in JSTS dataset.(3 times result and Average Acc.)

| Method | Accuracy 1 | Accuracy 2 | Accuracy 3 | Average Acc. |
|---|---|---|---|---|
| SLG 5-shot | **10.76** | **28.59** | **17.67** | **19.01** |
| SLG 10-shot | **17.19** | **16.95** | **20.00** | **18.05** |
| Simple Fine-tune 5-shot | 0 | 0 | 0 | 0 |
| Simple Fine-tune 10-shot | 0 | 0 | 0 | 0 |

As presented in Table 18, an examination of the JSTS dataset incorporating six classifications revealed that an elevated difficulty level correspondingly led to a reduction in accuracy. Notwithstanding this trend, the SLG framework managed to achieve accuracies of 19.01 and 18.05 in the 5-shot and 10-shot experiments, respectively. It was observed that the variability in the results was more pronounced in the 5-shot experiments, attributable to the smaller sample size. Upon increasing the sample size to 10, a significant reduction in the fluctuation of the results was notable.

In summation, the derived experimental findings corroborate the high proficiency of the SLG framework in conducting few-shot learning on a singular SC task dataset. This underlines its superior performance in this specific context.

Table 19  The experiment setup of few-shot experiment.

| Parameter | SLG framework | Simple fine-tune |
|---|---|---|
| Epoch | 5 | 5 |
| Batch Size | 8 | 8 |
| Learning Rate | 1e-4 | 1e-4 |
| Max Source Length | 256 | 256 |
| Max Source Length | 8 | 8 |
| Max Generate Length | 400 | 400 |

## 8 Conclusion and Future Work

In this study, we integrate Sentence Classification (SC) and Named Entity Recognition (NER) tasks, leveraging their shared knowledge to enhance accuracy. We propose the SCNM task and construct a comprehensive dataset from Wikipedia. Our experiments demonstrate mutual reinforcement effects between SC and NER, and introduce the versatile Sentence-to-Label Generate (SLG) framework for handling both tasks concurrently and individually through a Format Converter. In the later experiments, we modified FC to apply the SLG framework to two additional separate SC tasks. And we achieved far better performance than fine-tune in a series of experiments.

Future work includes exploring alternative language models, assessing the SLG framework on other individual NER datasets. And creating domain specific SCNM datasets to evaluate the SLG framework's adaptability and effectiveness.

## Acknowledgement

The authors would like to gratefully acknowledge the reviewers for their time and valuable comments. This research was partially supported by JSPS KAKENHI Grant Numbers JP23H00491 and JP22K00502. This paper is an extended version of paper (Gan et al. 2023) accepted for publication at the International Conference on Natural Language and Information Systems (NLDB 2023). This extended version contains the following content:
1. Added BART-base/large model as the base model of SLG framework. And we have conducted a comprehensive experiment on it.
2. A more complete ablation experiment is conducted to prove the effectiveness of SLG framework.
3. Added a fixed random seed experiment part. to more precisely control the variables to demonstrate the mutual reinforcement effect between SC and NER tasks.
4. The format converter was modified to apply the SLG framework to a separate SC task. A series of ablation experiments and few-shot learning experiments were also conducted.

**Chengguang Gan**: He is currently Ph.D student at Graduate School of Environment and Information Sciences, Yokohama National University. Before that, he graduated from The Kyoto College of Graduate Studies for Informatics in 2022. Before that, he graduated from East China Jiaotong University in 2018. His research interests are NLP, especially prompt method of language model.

**Qinghao Zhang**: has been currently pursuing the M.S. degree from the Department of Information Convergence Engineering, Pusan National University, Busan, South Korea, since 2021. His current research interests include natural language processing, artificial intelligence, and its applications.

**Tatsunori Mori**: received his B.E. degree in Information Engineering and Ph.D. degree in Electrical and Computer Engineering from Yokohama National University in 1986 and 1991, respectively. After being a research associate, he is currently professor in the Graduate School of Environment and Information Sciences, Yokohama National University. He was a visiting scholar in the Center for the Study of Language and Information (CSLI), Stanford University, USA, from February to November 1998. He is a member of the Association for Natural Language Processing, Information Processing Society of Japan, the Institute of Electronics, Information and Communication Engineers, the Japanese Society for Artificial Intelligence, and ACM.